\documentclass{Interspeech}

\interspeechcameraready

\usepackage{algorithmic}
\usepackage{bm}
\usepackage{subcaption}
\usepackage{comment}
\usepackage{cite}
\usepackage{multirow}

\title{DYNAC: Dynamic Vocabulary-based Non-Autoregressive Contextualization for Speech Recognition}

\author[affiliation={1}]{Yui}{Sudo}
\author[affiliation={1}]{Yosuke}{Fukumoto}
\author[affiliation={1}]{Muhammad}{Shakeel}
\author[affiliation={2}]{Yifan}{Peng}
\author[affiliation={2}]{Chyi-Jiunn}{Lin}
\author[affiliation={2}]{Shinji}{Watanabe}

\affiliation{}{Honda Research Institute Japan}{Japan}
\affiliation{}{Carnegie Mellon University}{USA}
\email{shakeel.muhammad@jp.honda-ri.com, yifanpen@andrew.cmu.edu, shinjiw@ieee.org}

\usepackage{comment}

\begin{document}

\maketitle

\begin{abstract} 
Contextual biasing (CB) improves automatic speech recognition for rare and unseen phrases. Recent studies have introduced dynamic vocabulary, which represents context phrases as expandable tokens in autoregressive (AR) models. This method improves CB accuracy but with slow inference speed. While dynamic vocabulary can be applied to non-autoregressive (NAR) models, such as connectionist temporal classification (CTC), the conditional independence assumption fails to capture dependencies between static and dynamic tokens. This paper proposes DYNAC (Dynamic Vocabulary-based NAR Contextualization), a self-conditioned CTC method that integrates dynamic vocabulary into intermediate layers. Conditioning the encoder on dynamic vocabulary, DYNAC effectively captures dependencies between static and dynamic tokens while reducing the real-time factor (RTF). Experimental results show that DYNAC reduces RTF by 81\% with a 0.1-point degradation in word error rate on the LibriSpeech 960 test-clean set.
\end{abstract}

\section{Introduction}

End-to-end automatic speech recognition (E2E-ASR) \cite{prabhavalkar2023end,li2022recent} can be classified into autoregressive (AR) and non-autoregressive (NAR) models. AR models, such as the attention-based encoder-decoder \cite{chorowski2015attention,attention1,radford2023robust,
peng2024owsm} and recurrent neural network transducer (RNN-T) \cite{rnnt1,gulati2020conformer}, predict the next token based on the previous token sequence in an AR manner. 
In contrast, NAR models, such as connectionist temporal classification (CTC) \cite{ctc1,ctc2,peng2024owsmctc,Higuchi2021ACS}, predict all tokens simultaneously under the conditional independence assumption. This approach allows faster inference but often results in lower performance compared to AR models. 
Several methods have been proposed to relax the conditional independence assumption \cite{lee2021intermediate,Chan2020ImputerSM,higuchi2020mask,chi-etal-2021-align,9052964}. For example, self-conditioned CTC incorporates intermediate predictions as feedback to refine subsequent encoder layers, allowing the model to capture contextual dependencies \cite{Nozaki2021RelaxingTC}. 
Despite these advances, the performance of E2E-ASR models remains highly dependent on the training data, leading to performance inconsistencies for rare and unseen phrases. 
Frequent retraining to adapt to these phrases is impractical, highlighting the need for a method to contextualize models without additional retraining.

Contextual biasing (CB) \cite{deepcontext2018,Jain2020ContextualRF,huber2021instant,zhou2023copyne,sudo2024_bias} provides an effective strategy for adapting E2E-ASR models to specific context phrases through a bias list without requiring additional retraining. Most CB methods are designed for AR models, leveraging a cross-attention layer in the decoder for accurate bias phrase recognition. Although AR-CB methods achieve accurate bias phrase recognition, their slow inference speed makes them unsuitable for latency-sensitive applications.
Several CB methods have been proposed for NAR models or adapted to CTC-based NAR models to improve inference speed \cite{huang2023contextualized,shakeel2024_bias,nakagome24_interspeech}. 
However, most existing CB methods represent bias phrases as sequences of subwords from a pre-defined vocabulary (referred to as static vocabulary), resulting in unnatural token patterns with low occurrence probabilities. 
For example, the personal name ``\textit{Raphael}'' might be segmented into ``\textit{Ra}'', ``\textit{pha}'', and ``\textit{el}''. If such static token patterns are rare in the training data, their recognition accuracy decreases significantly.
Several studies mitigate this issue by employing additional information, such as text-only data \cite{Le2021ContextualizedSE,qiu2023improving}, synthesized speech \cite{wang2022towards,wang2023improving}, phonemes \cite{bruguier2019phoebe,futami2023phoneme}, 
and named entity tags \cite{sudo2023retraining}. While these approaches improve CB performance, they introduce additional computational and operational overhead.

To mitigate this issue without requiring such additional information, dynamic vocabulary expansion has been proposed \cite{sudo2024contextualized}. This method introduces a dynamic vocabulary, where each bias phrase is represented as a dynamically expandable single token instead of being decomposed into a sequence of static tokens. For example, the bias phrase ``\textit{Raphael}” is represented as a single dynamic token [$<$\textit{Raphael}$>$] rather than the static token sequence [``\textit{Ra}'', ``\textit{pha}'', ``\textit{el}''].
While this method improves CB performance without relying on additional information, its evaluation has been limited to AR models (CTC/attention and CTC/RNN-T) \cite{watanabe2017hybrid,sudo20244d}, which are computationally expensive. 
Although dynamic vocabulary expansion can be applied to CTC-based NAR models, the conditional independence assumption limits the ability to capture contextual dependencies between static and dynamic tokens.

This paper proposes DYNAC (Dynamic Vocabulary-based NAR Contextualization), an NAR-CB method using self-conditioned CTC that incorporates dynamic vocabulary into the intermediate encoder layers.
By conditioning the encoder on the dynamic vocabulary, DYNAC relaxes the conditional independence assumption. This approach allows the model to capture dependencies between static and dynamic tokens while reducing the real-time factor (RTF) compared to AR-CB methods \cite{sudo2024contextualized}. 
The main contributions of this paper are as follows:
\begin{itemize}
    \item We propose DYNAC, an NAR-CB method using self-conditioned CTC that integrates dynamic vocabulary, enabling efficient CB while reducing the RTF.
    \item We analyze the effectiveness of DYNAC by comparing the token-wise scores of the static and dynamic vocabulary to highlight its impact on recognition accuracy.
    \item We demonstrate the effectiveness of DYNAC on the LibriSpeech 960 corpus and our in-house Japanese dataset, showing robust performance even on unseen phrases.
\end{itemize}

\section{CTC-based CB with dynamic vocabulary}
\label{sec:Preliminary}

This section describes the CTC-based CB method using dynamic vocabulary, which consists of an audio encoder, a bias encoder, and a scoring layer based on CTC. Although dynamic vocabulary expansion was originally proposed for AR-CB methods, it can be applied to CTC-based NAR models \cite{sudo2024contextualized}.

\subsection{Audio encoder}
\label{sec:encoder}

We use the Conformer \cite{gulati2020conformer} for the audio encoder, comprising $L$ encoder layers.
The $l$-th encoder layer transforms its $T$-length $d$-dimensional hidden state representation input $\bm{X}^{\text{in}}_{(l)} \in \mathbb{R}^{T \times d}$ into $\bm{X}^{\text{out}}_{(l)} \in \mathbb{R}^{T \times d}$ as follows:
\begin{equation}
\label{eq:audio}
    \bm{X}^{\text{out}}_{(l)} = \mathrm{AudioEnc}_{(l)}(\bm{X}^{\text{in}}_{(l)}).
\end{equation}
\vspace*{-1mm}
Then, the output of the $l$-th encoder layer $\bm{X}^{\text{out}}_{(l)}$ is directly used as the input of the ($l$+1)-th encoder layer $\bm{X}^{\text{in}}_{(l + 1)}$ as follows:
\vspace*{-1mm}
\begin{equation}
\label{eq:inter_layer}
    \bm{X}^{\text{in}}_{(l + 1)} = \bm{X}^{\text{out}}_{(l)}.
\end{equation}
By iterating this process $L$ times, the final output $\bm{X}^{\text{out}}_{(L)} \in \mathbb{R}^{T \times d}$ is obtained.

\subsection{Bias encoder}
\label{sec:biasencoder}

The bias encoder comprises Transformer \cite{NIPS2017_3f5ee243} layers and a mean pooling layer with a bias list $\bm{B} = \{b_{1}, \cdots, b_{N}$\}, where $N$ represents the number of bias phrases.
Each bias phrase $b_{n}$ is represented as a sequence of static tokens (e.g., [``\textit{Ra}'', ``\textit{pha}'', ``\textit{el}'']) in the pre-defined static vocabulary $\mathcal{V}^{\text{static}}$.
The bias list $\bm{B}$ is processed by the bias encoder to obtain phrase-level representations $\bm{V} = [\bm{v}_1, \cdots , \bm{v}_{N}] \in \mathbb{R}^{N \times d}$ as follows:
\vspace*{-1mm}
\begin{equation}
    \bm{V} = \mathrm{BiasEnc}(\bm{B}).
\label{eq:hp}
\end{equation}
The bias phrases in the bias list $\bm{B} = \{b_{1}, \cdots, b_{N}$\} are incorporated into the dynamic vocabulary $\mathcal{V}^{\text{dynamic}} = \{<\hspace{-3pt}b_1\hspace{-3pt}>, \cdots, <\hspace{-3pt}b_N\hspace{-3pt}>\}$ of size $N$, where each entry represents the corresponding bias phrase as a single dynamic token (e.g., [$<$\textit{Raphael}$>$]).

\subsection{CTC scoring layer with dynamic vocabulary}
\label{sec:ctc}

Similar to conventional CTC, the scoring layer predicts an alignment sequence $A = [a_1, \dots, a_T]$ instead of directly predicting the output token sequence $Y$. Unlike conventional CTC, the scoring layer expands the static vocabulary $\{\mathcal{V}^{\text{static}} \cup \{\phi\}\}$ of size $K$ to include the dynamic vocabulary $\mathcal{V}^{\text{dynamic}}$ of size $N$ by incorporating $\bm{V}$ as follows:
\vspace*{-1mm}
\begin{equation}
\label{eq:ctc-alignment}
    P(Y \mid \bm{X}^{\text{out}}_{(L)}, \bm{V}) = \sum_{A \in \mathcal{B^{\text{-1}}}(Y)} P(A \mid \bm{X}^{\text{out}}_{(L)}, \bm{V}),
\end{equation}
where $\phi$ represents the blank token, and $\mathcal{B^{\text{-1}}}(Y)$ represents the set of all possible alignment sequences of token sequence $Y$.
$P(A | \bm{X}^{\text{out}}_{(L)}, \bm{V})$ is computed based on the conditional independence assumption as follows:
\vspace*{-2mm}
\begin{equation}
\label{eq:ctc}
    P(A \mid \bm{X}^{\text{out}}_{(L)}, \bm{V}) = \prod_{t=1}^{T} P\left(a_{t} \mid \bm{X}^{\text{out}}_{(L)}, \bm{V} \right).
\end{equation}
To estimate the alignment probability in Eq. \eqref{eq:ctc},
the scoring layer is formulated as follows:
\vspace*{-1mm}
\begin{align}
    \label{eq:s_score}
    & \bm{S}^{\text{static}}_{(L)} = \mathrm{Linear}(\bm{X}^{\text{out}}_{(L)}), \\
    \label{eq:d_score}
    & \bm{S}^{\text{dynamic}}_{(L)} = \frac{\mathrm{Linear}(\bm{X}^{\text{out}}_{(L)}) \mathrm{Linear}(\bm{V}^{T})}{\sqrt{d}}, \\
    \label{eq:prob}
    & \bm{Z}_{(L)} = \mathrm{Softmax}(\mathrm{Concat}(\bm{S}^{\text{static}}_{(L)}, \bm{S}^{\text{dynamic}}_{(L)})), 
\end{align}
where $\bm{S}^{\text{static}}_{(L)} \in \mathbb{R}^{T \times K}$ and $\bm{S}^{\text{dynamic}}_{(L)} \in \mathbb{R}^{T \times N}$ represent the alignment scores for static and dynamic vocabulary, respectively.
$\bm{Z}_{(L)} 
\in \mathbb{R}^{T \times (K + N)}$ denotes the alignment probability distribution combining static and dynamic vocabulary. $\bm{Z}_{(L)}$ is used to predict the alignment probability $P(A \mid \bm{X}^{\text{out}}_{(L)}, \bm{V} )$ in Eq. \eqref{eq:ctc}.
The model parameters are optimized by minimizing the negative log-likelihood as follows:
\vspace*{-1mm}
\begin{equation}
    L_{\text{ctc}} = - \log P(Y \mid \bm{X}^{\text{out}}_{(L)}, \bm{V}) = - \log P(Y \mid \bm{Z}_{(L)}).
\label{eq:loss_ctc}
\end{equation}

While CTC's conditional independence assumption in Eq. \eqref{eq:ctc} allows fast inference, it often leads to suboptimal performance. In particular, since the dynamic vocabulary is incorporated only in the final layer in Eqs. \eqref{eq:d_score} and \eqref{eq:prob}, the model cannot capture dependencies between dynamic tokens and surrounding static tokens, which results in poor recognition accuracy.

\section{DYNAC}
\label{sec:proposed}

\begin{figure}[t]
        \begin{minipage}{0.49\textwidth}
        \centering
            \includegraphics[width=0.95\textwidth]{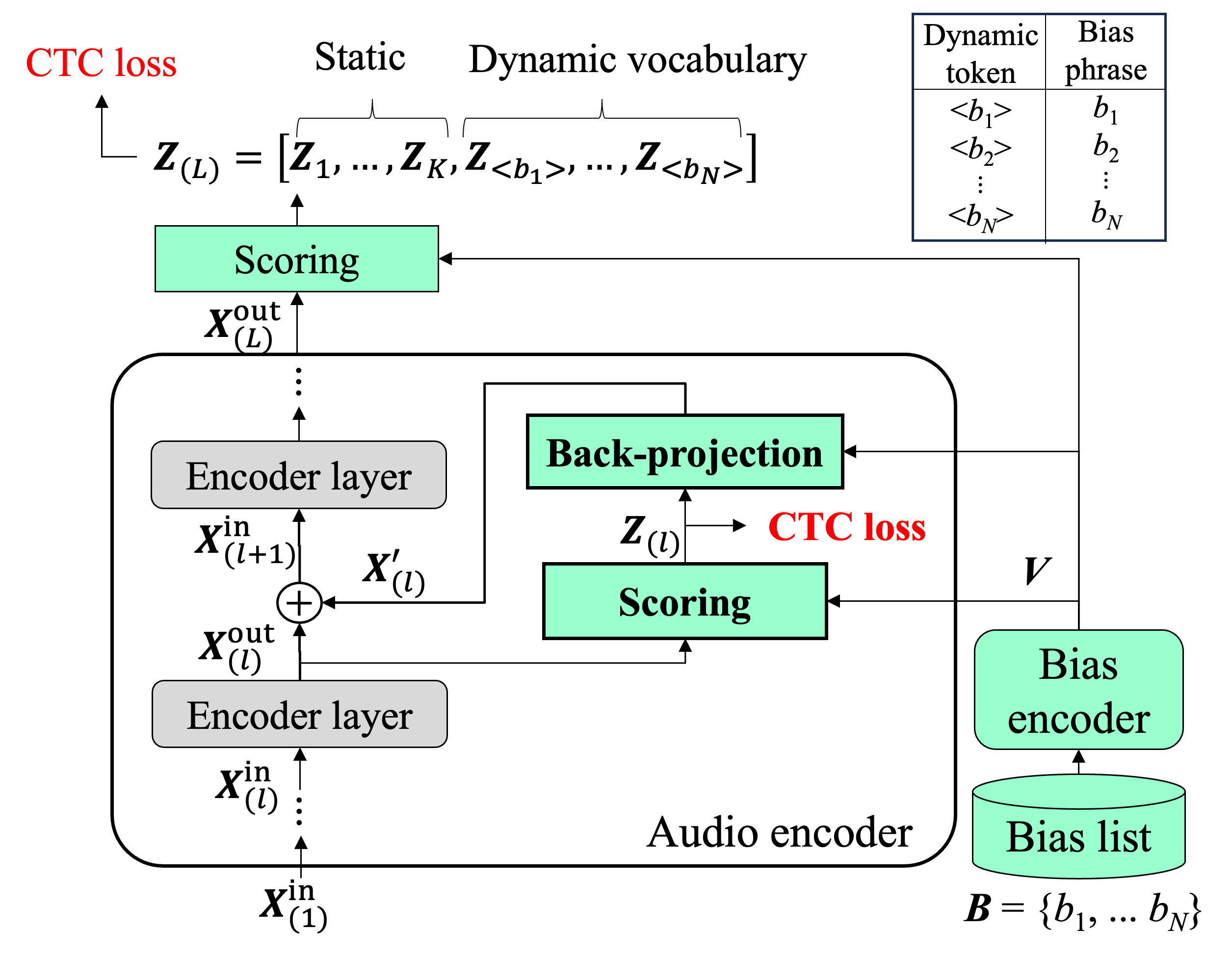} 
        \end{minipage}
    \vspace*{-3mm}
    \caption{Overall architecture of DYNAC. Bolded blocks highlight the expansions beyond the conventional CTC-based CB method with dynamic vocabulary. 
    }
    \label{fig:overview}
\vspace*{-0mm}
\end{figure}

To address the limitation described in Section~\ref{sec:ctc}, we introduce scoring and back-projection layers into the intermediate encoder layers. 
This architecture extends the conventional self-conditioned CTC~\cite{Nozaki2021RelaxingTC}, enabling the integration of dynamic vocabulary into intermediate representations of the encoder, as shown in Figure~\ref{fig:overview}.

\begin{figure}[t]
     \centering
     \hfill
     \begin{subfigure}[b]{0.49\linewidth}
         \centering
         \includegraphics[scale=0.35]{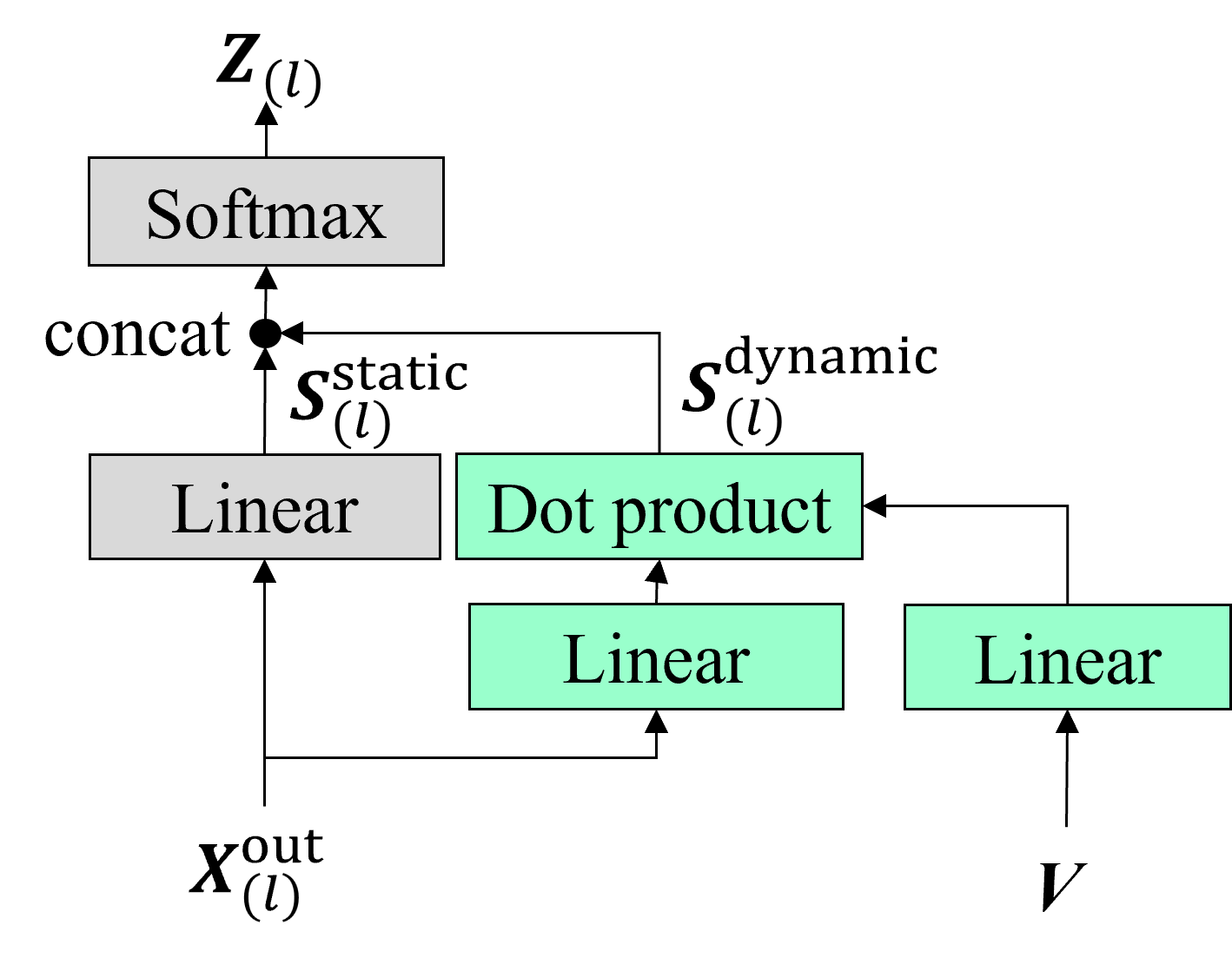}
         \vskip -2mm
         \caption{Scoring layer}
         \label{fig:scoring}
     \end{subfigure}
     \hfill
     \begin{subfigure}[b]{0.49\linewidth}
         \centering
         \includegraphics[scale=0.35]{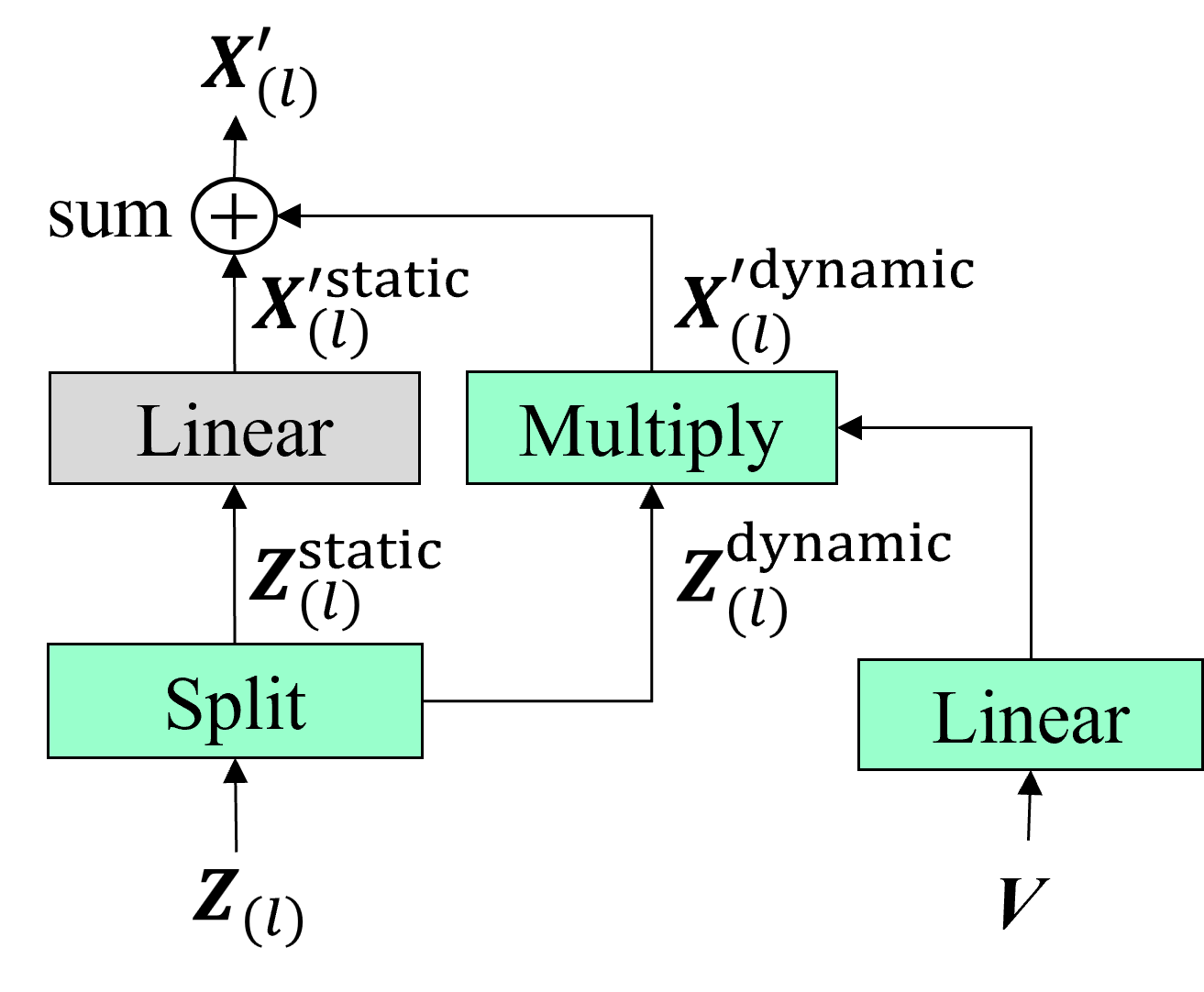}
         \vskip -2mm
         \caption{Back-projection layer }
         \label{fig:vectorize}
     \end{subfigure}
     \hfill
    \vskip -0.1in
    \caption{Components of DYNAC.}
    \label{fig:components}
    \vskip -1mm
\end{figure}

\subsection{Self-conditioned CTC with dynamic vocabulary}
\label{sec:inter-scoring}
\vspace*{-0mm}

The scoring layer estimates the token-wise probability $\bm{Z}_{(l)} \in \mathbb{R}^{T \times (K + N)}$ similar to Eqs. \eqref{eq:s_score}, \eqref{eq:d_score}, and \eqref{eq:prob}, but using the output of the $l$-th intermediate layer $\bm{X}^{\text{out}}_{(l)}$ and $\bm{V}$ (Figure \ref{fig:scoring}).

Then, the back-projection layer (Figure \ref{fig:vectorize}) converts the token-wise probability $\bm{Z}_{(l)}$ back into the $d$-dimensional hidden representation $\bm{X}^{\prime}_{(l)} \in \mathbb{R}^{T \times d}$ while maintaining the dynamic vocabulary information.
Since $\bm{Z}_{(l)}$ contains both static and dynamic vocabulary, which have fixed and dynamically changing sizes, we first split $\bm{Z}_{(l)}$ into the probabilities for the static and dynamic vocabulary components ($\bm{Z}^{\text{static}}_{(l)} \in \mathbb{R}^{T \times K}$ and $\bm{Z}^{\text{dynamic}}_{(l)} \in \mathbb{R}^{T \times N}$) as follows:
\begin{equation}
    \label{eq:split}
    \bm{Z}^{\text{static}}_{(l)}, \bm{Z}^{\text{dynamic}}_{(l)} = \mathrm{Split}(\bm{Z}_{(l)}).
\end{equation}
Subsequently, $\bm{Z}^{\text{static}}_{(l)}$ and $\bm{Z}^{\text{dynamic}}_{(l)}$ are transformed to the $d$-dimensional hidden representation $\bm{X}^{\prime}_{(l)}$ as follows:
\vspace*{-1mm}
\begin{equation}
    \label{eq:vectorize}
    \bm{X}^{\prime}_{(l)} = \mathrm{Linear}(\bm{Z}^{\text{static}}_{(l)}) + \bm{Z}^{\text{dynamic}}_{(l)} \bm{V}.
\end{equation}
Here, the $T$-length $d$-dimensional hidden representation is obtained by multiplying $\bm{Z}^{\text{dynamic}}_{(l)} \in \mathbb{R}^{T \times N}$ with the bias phrase representation $\bm{V} \in \mathbb{R}^{N \times d}$ in Eq. \eqref{eq:hp}. This transformation does not contain trainable parameters, allowing for efficient handling of dynamically changing vocabulary.
Finally, the hidden representation $\bm{X}^{\prime}_{(l)}$ is added to $\bm{X}^{\text{out}}_{(l)}$, unlike Eq. \eqref{eq:inter_layer}, and used as the input of the ($l$+1)-th layer as follows:
\begin{equation}
    \bm{X}^{\text{in}}_{(l + 1)} = \bm{X}^{\text{out}}_{(l)} + \bm{X}^{\prime}_{(l)}.
\end{equation}
This feedback mechanism (Figure \ref{fig:overview}) enables the encoder to refine feature representations by incorporating contextual information through self-attention layers in the encoder, thereby capturing dependencies between static and dynamic tokens.

\subsection{Intermediate CTC loss}
\label{sec:inter-ctc}
\vspace*{-0mm}

To train DYNAC effectively, we follow \cite{Nozaki2021RelaxingTC} and introduce an auxiliary CTC loss at intermediate layers:
\begin{equation}
    L_{\text{inter}} = - \dfrac{1}{|\mathcal{S}|} \sum_{l \in \mathcal{S}} \log P(Y \mid \bm{Z}_{(l)}).
\label{eq:interctc_loss}
\end{equation}
$\mathcal{S}$ represents the set of intermediate layers where the auxiliary loss is applied.
We also introduce an auxiliary attention loss during training following \cite{sudo2024contextualized}. 
The model parameters are optimized by combining Eqs. \eqref{eq:loss_ctc} and \eqref{eq:interctc_loss} using a tunable hyper-parameter $\lambda$ as follows:
\begin{equation}
    L_{\text{total}} = \lambda L_{\text{ctc}} + \lambda L_{\text{inter}} + (1 - 2 \lambda) L_{\text{att}}.
\label{eq:total_loss}
\end{equation}

\subsection{Training and inference}
\label{sec:training}

We follow the same training and inference strategy as \cite{sudo2024contextualized}. 
For each batch, a bias list $\bm{B}$ containing $N$ bias phrases is randomly generated from the reference transcriptions.
Once the bias list $\bm{B}$ is defined, the corresponding static tokens are replaced with the dynamic tokens.
For example, if [``\textit{Ra}'', ``\textit{pha}'', ``\textit{el}''] is extracted from a complete utterance [``\textit{Hi}'', ``\textit{Ra}'', ``\textit{pha}'', ``\textit{el}''], the reference transcription is modified to [``\textit{Hi}'', $<$\textit{Raphael}$>$].
During inference, we apply the bias weight to adjust the token-wise probability $\bm{Z}_{(L)}$ in Eq. \eqref{eq:prob} to avoid over/under-biasing. 
Specifically, we apply the tunable weight $\mu$ to the dynamic token probability ($\mu \bm{Z}^{\text{dynamic}}_{(L)}$).
If $\mu < 1.0$, the dynamic tokens are underweighted compared to the static tokens; otherwise, the dynamic tokens are overweighted compared to the static tokens.

\section{Experiment}

We conduct several experiments to verify the effectiveness of the proposed method.

\subsection{Experimental setup}
\label{sec:experimental condition}
\vspace*{-0mm}

The input features are 80-dimensional Mel filterbanks with a window size of 512 samples and a hop length of 160 samples. Then, SpecAugment \cite{specaug} is applied.
In DYNAC, the audio encoder comprises two convolutional layers with a stride of two and a 256-dimensional linear projection layer followed by 12 Conformer layers with 1024 linear units. 
The proposed self-conditioned CTC architecture is applied at the $\mathcal{S} = \{3,6,9\}$ intermediate layers (Section \ref{sec:inter-ctc}). 
The bias encoder has six Transformer blocks with 1024 linear units.
DYNAC has 41.92 M parameters, including the bias encoder.
During training, a bias list $\bm{B}$ is created randomly for each batch, resulting in a total of $N$ = 50 $\sim$ 200 bias phrases (Section~\ref{sec:training}). 
The proposed models are trained with a static vocabulary size $K$ of 5,000 for 70 epochs at a learning rate of 0.002 with 15,000 warmup steps using the Adam optimizer.
The training weight $\lambda$ in Eq. \eqref{eq:total_loss} and the bias weight $\mu$ (Section \ref{sec:training}) are set to 0.15 and 0.1, respectively. 

The proposed method is evaluated on the LibriSpeech-960 corpus \cite{panayotov2015librispeech} using the ESPnet toolkit \cite{espnet}, in terms of overall word error rate (WER), biased phrase WER (B-WER), and unbiased phrase WER (U-WER) \cite{Le2021ContextualizedSE}.
In addition, we measure RTF using a CPU (Intel(R) Xeon(R) Platinum 8480C @2.00GHz) to assess inference speed.
Note that the bias encoder is excluded from RTF measurement because it is only executed once when the bias list is updated (e.g., once before a session), and the resulting $\bm{V}$ in Eq. \eqref{eq:hp} is reused for subsequent use.
The goal of this study is to improve B-WER and RTF while minimizing degradation in U-WER. 

\subsection{Main results}
\label{sec:main results}
\vspace*{-0mm}

Table \ref{main_table} presents the results of WER and RTF on the LibriSpeech 960 test-clean set with a bias list size of $N$ = 1000. To evaluate the impact of inference speed reduction compared to AR models, we include results from the CTC/attention model and the AR-CB method with dynamic vocabulary \cite{sudo2024contextualized} with a beam size of 5. For NAR models, we report results from the conventional self-conditioned CTC \cite{Nozaki2021RelaxingTC} and the CTC-based CB method with dynamic vocabulary (Section \ref{sec:Preliminary}).

DYNAC significantly reduces RTF while maintaining a WER comparable to the AR-CB method (A3 vs. B3). Compared to the NAR baseline, DYNAC slightly increases the RTF due to the introduction of the proposed self-conditioned CTC architecture, but substantially improves B-WER (B1 vs. B3). Notably, while applying the dynamic vocabulary-based CB method to a CTC-based NAR model (Section \ref{sec:Preliminary}) improves B-WER, it severely degrades U-WER (B2). This issue is further analyzed in Section \ref{sec:analysis}.

\begin{table}[t]
\caption{Comparison between AR and NAR models on the LibriSpeech 960 test-clean set. \textbf{Bold} values represent the best results among the same category.
}
\vspace*{-7mm}
\label{main_table}
\begin{center}
\resizebox {0.99\linewidth} {!} {
\begin{tabular}{@{}clcccc}
\hline
ID & Model & WER & U-WER & B-WER & RTF \\
\hline
 \multicolumn{3}{l}{\textbf{\textit{Autoregressive}}} \\
A1 & CTC/attention             & 3.0 & \textbf{1.9} & 12.3 & 0.213 \\ 
A2 & BPB beam search \cite{sudo2024_bias} & 3.5 & 3.0 & 7.7 & N/A\\
A3 & Dynamic vocabulary \cite{sudo2024contextualized} & \textbf{2.0} & \textbf{1.9} & \textbf{3.3} & \textbf{0.165 }\\
\hline
\multicolumn{3}{l}{\textbf{\textit{Non-autoregressive}}} \\
B1 & Self-conditioned CTC \cite{Nozaki2021RelaxingTC} & 3.1 & \textbf{1.8} & 14.1 & 0.027 \\ 
B2 & CTC-based CB (Section \ref{sec:Preliminary}) & 40.8 & 45.2 & 6.9 & \textbf{0.024} \\
B3 & DYNAC (ours) & \textbf{2.1} & 1.9 & \textbf{3.2} & 0.031 \\
\hline
\end{tabular}
}
\end{center}
\vspace*{-7mm}
\end{table}

\begin{table*}[t]
\caption{WER results on the Librispeech-960 (U-WER/B-WER). \textbf{Bold} values represent the best result among the same bias list size $N$.}
\vspace*{-7mm}
\label{scalability}
\begin{center}
\resizebox {\linewidth} {!} {
\begin{tabular}{@{}c|cccc|cccc}
\hline
      & \multicolumn{4}{c|}{test-clean} & \multicolumn{4}{c}{test-other} \\
Model & $N$=0 (no-bias) & $N$=100 & $N$=500 & $N$=1000 & $N$=0 (no-bias) & $N$=100 & $N$=500 & $N$=1000 \\
\hline
\multirow{2}{*}{Baseline (CTC)} & 3.5 & 3.5 & 3.5 & 3.5 & 8.2 & 8.2 & 8.2 & 8.2\\
& (1.9/16.2) & (1.9/16.2) & (1.9/16.2) & (1.9/16.2) & (5.4/33.7) & (5.4/33.7) & (5.4/33.7) & (5.4/33.7)\\
\hline
\multirow{2}{*}{Self-cond. CTC \cite{Nozaki2021RelaxingTC}} & 3.1 & 3.1 & 3.1 & 3.1 &\textbf{7.3} & 7.3 & 7.3 & 7.3 \\
& (1.8/14.1) & (1.8/14.1) & (\textbf{1.8}/14.1) & (\textbf{1.8}/14.1) & (\textbf{4.7}/30.2) & (\textbf{4.7}/30.2) & (\textbf{4.7}/30.2) & (\textbf{4.7}/30.2)\\
\hline
\multirow{2}{*}{CTC-based CPPNet \cite{huang2023contextualized}} & 4.0 & 3.8 & 3.8 & 3.9 & 9.0 & 8.6 & 8.7 & 8.9 \\
& (2.3/17.9) & (2.2/17.0) & (2.1/17.5) & (2.4/18.3) & (5.9/35.7) & (5.6/34.4) & (5.6/35.73) & (5.7/37.3) \\
\hline
\multirow{2}{*}{Intermediate CB \cite{shakeel2024_bias}} & \textbf{3.0} & 2.9 & 3.1 & 3.3 & 7.4 & 7.2 & 7.7 & 8.2 \\
& (\textbf{1.7}/\textbf{13.0}) & (\textbf{1.6}/12.8) & (\textbf{1.8}/13.4) & (1.9/14.4) & (5.0/\textbf{28.6}) & (\textbf{4.8}/28.1) & (5.1/30.7) & (5.5/32.2) \\
 \hline
\multirow{2}{*}{DYNAC (ours)} & 3.5 & \textbf{2.0} & \textbf{1.9} & \textbf{2.1} & 8.5 & \textbf{5.4} & \textbf{5.4} & \textbf{5.8} \\
 & (2.0/16.2) & (1.8/\textbf{4.2}) & (\textbf{1.8}/\textbf{3.0}) & (1.9/\textbf{3.2}) & (5.4/35.4) & (4.9/\textbf{10.0}) & (5.0/\textbf{8.3}) & (5.4/\textbf{9.0}) \\
\hline
\end{tabular}
}
\end{center}
\vspace*{-5mm}
\end{table*}

\subsection{Analysis on token-wise scores}
\label{sec:analysis}
\vspace*{-0mm}

\begin{figure}[t]
     \centering
     \hfill
     \begin{subfigure}[b]{0.49\linewidth}
         \centering
         \includegraphics[scale=0.26]{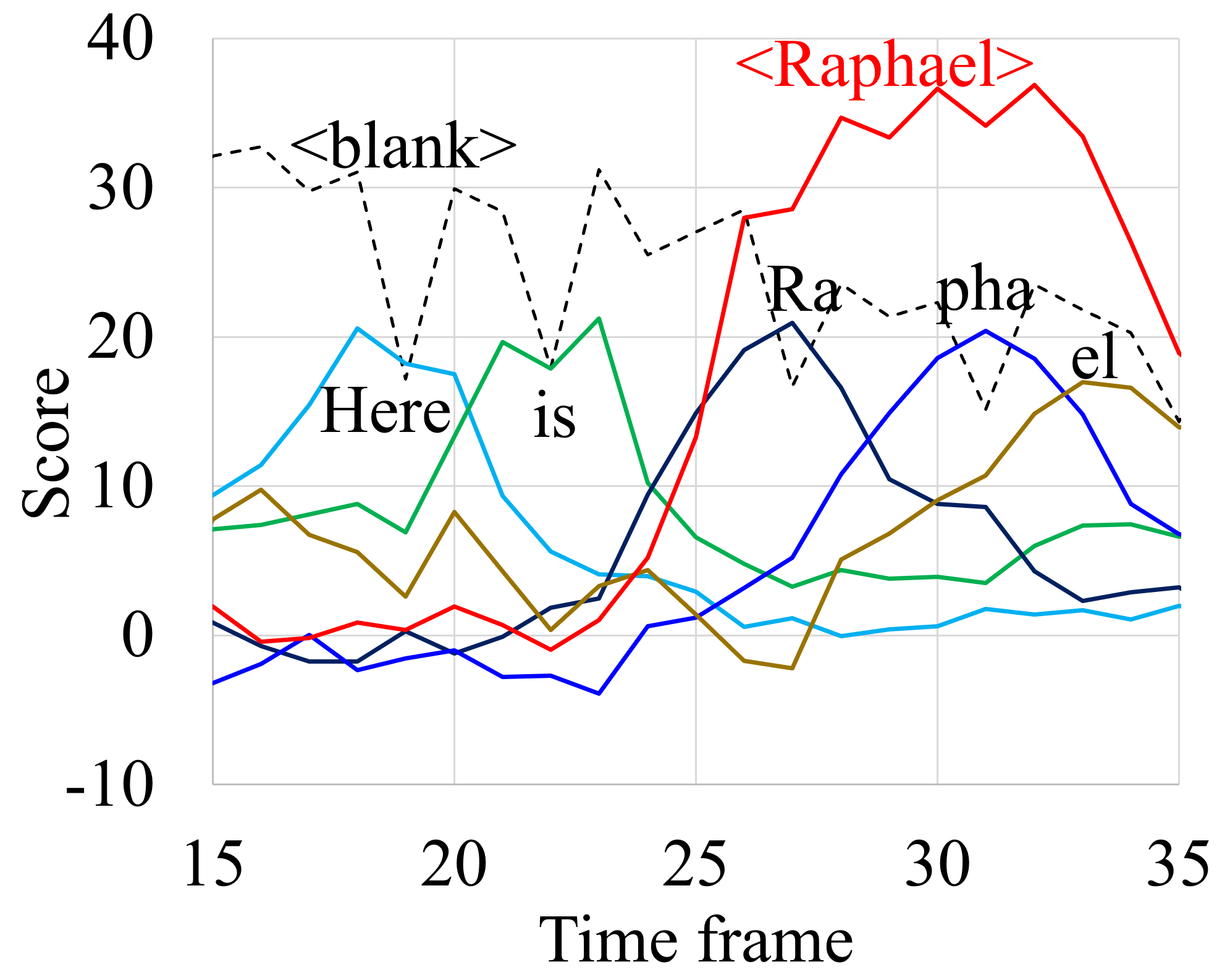}
         \vskip -1.0mm
         \caption{CTC-based CB (Section \ref{sec:Preliminary})}
         \label{fig:ctc_score}
     \end{subfigure}
     \hfill
     \begin{subfigure}[b]{0.49\linewidth}
         \centering
         \includegraphics[scale=0.26]{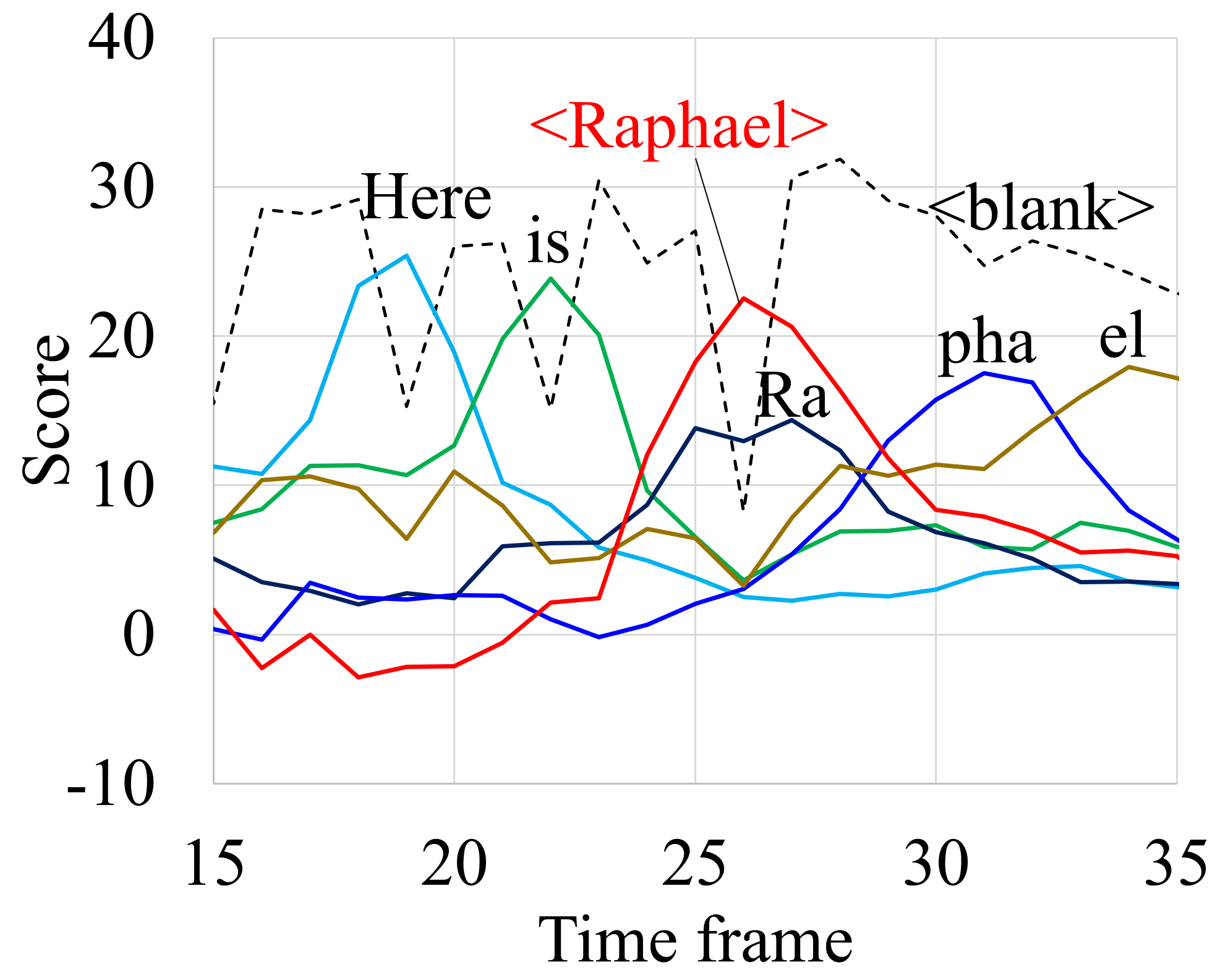}
         \vskip -1.0mm
         \caption{DYNAC}
         \label{fig:sc_score}
     \end{subfigure}
     \hfill
    \vskip -3mm
    \caption{Comparison in token-wise score.}
    \label{fig:score}
    \vskip -3mm
\end{figure}

Figure \ref{fig:score} compares the token-wise scores ($\bm{S}^{\text{static}}_{(L)}$ and $\bm{S}^{\text{dynamic}}_{(L)}$ in Eqs. \eqref{eq:s_score} and \eqref{eq:d_score}) between the conventional CTC-based CB method (Section \ref{sec:Preliminary}) and DYNAC.
The red line represents the score of the dynamically expanded token $<$\textit{Raphael}$>$, while the dashed and other lines show the blank and static token scores, respectively.

In the CTC-based CB method (Figure \ref{fig:ctc_score}), dynamic vocabulary is incorporated only in the final layer, preventing dependencies between static and dynamic tokens from being captured. As a result, the dynamic token $<$\textit{Raphael}$>$ receives an excessively high score, while static tokens (``\textit{Here}'', ``\textit{is}'') have lower scores than blank tokens, leading to a substantial degradation in U-WER (Table \ref{main_table}).
In contrast, DYNAC (Figure \ref{fig:sc_score}) addresses this issue by introducing self-conditioned CTC, allowing intermediate layers to integrate dynamic vocabulary. This approach ensures that static token scores remain higher than blank scores while preventing excessive emphasis on dynamic tokens, thereby mitigating U-WER degradation.

\subsection{Impact of bias list size}
\label{sec:scalability}
\vspace*{-0mm}

Table \ref{scalability} presents the impact of the bias list size $N$ and compares DYNAC with existing NAR-CB methods. While the existing CB methods \cite{huang2023contextualized,shakeel2024_bias} were primarily designed for AR models, we apply them to CTC-based NAR models for a fair comparison.
DYNAC consistently improves B-WER significantly even as the bias list size increases, resulting in a better overall WER than the baseline. Moreover, DYNAC outperforms existing CB methods by a large margin.

\subsection{Effectiveness on rare and unseen phrases}
\label{sec:frequency}

\begin{figure}[t]
        \begin{minipage}{0.49\textwidth}
        \centering
            \includegraphics[width=0.7\textwidth]{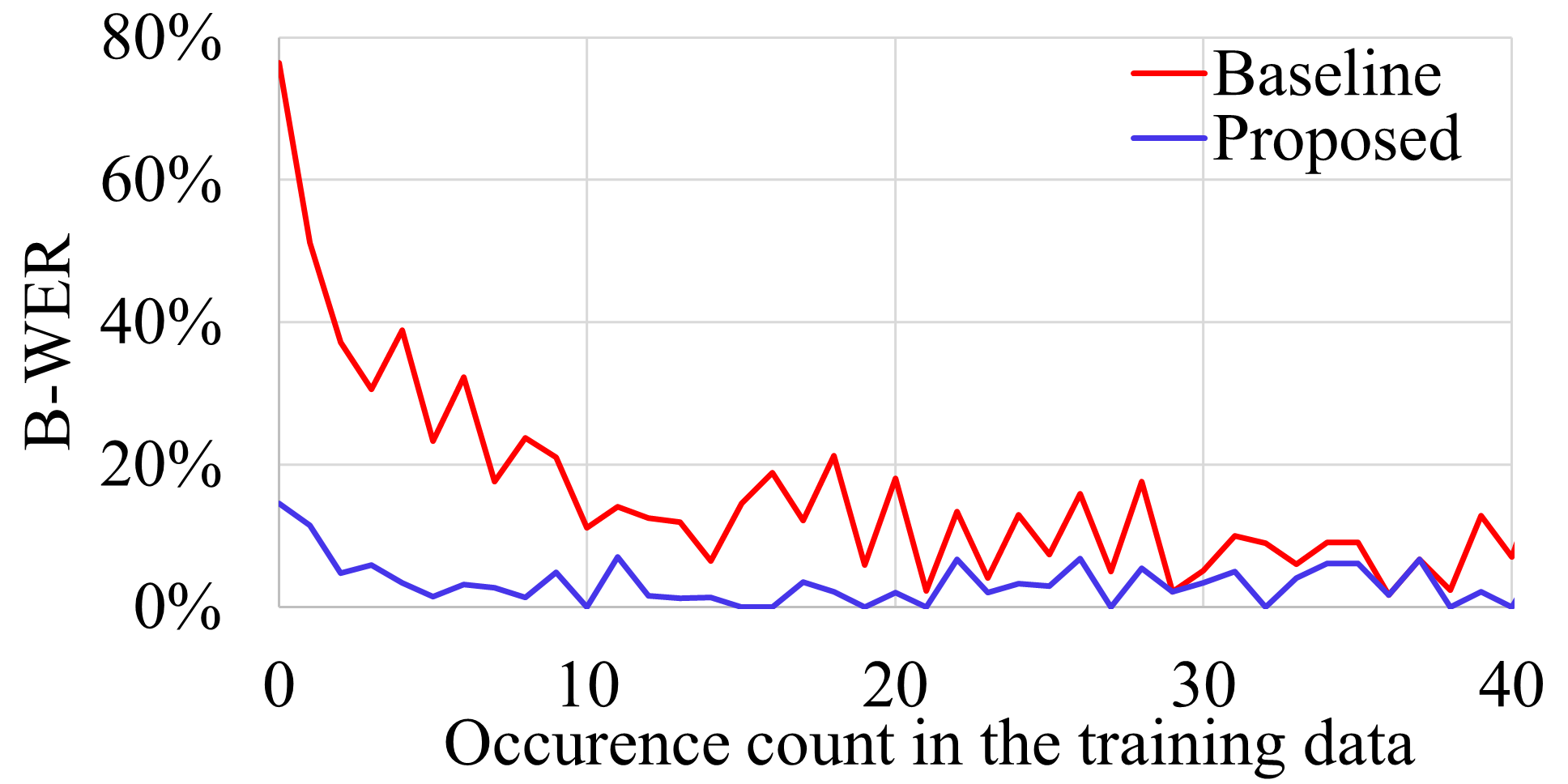} 
        \end{minipage}
    \vspace*{-3mm}
    \caption{Performance evaluation on rare and unseen phrases.}
    \label{fig:unseen}
\vspace*{-0mm}
\end{figure}

Figure \ref{fig:unseen} illustrates the relationship between the phrase occurrence in the training data and the B-WER. The red and blue lines represent the baseline self-conditioned CTC \cite{Nozaki2021RelaxingTC} and DYNAC with a bias list size of $N = 1000$, respectively. Phrases that do not occur in the training data (zero occurrences) correspond to unseen words.
In the baseline model, B-WER increases as phrase occurrence decreases, reaching 76\% for unseen words. In contrast, DYNAC consistently improves B-WER across all occurrence count ranges, with particularly notable gains for words occuring less than 20 times. Moreover, DYNAC remains effective even for unseen phrases, achieving a B-WER of 14.6\%.

\subsection{Validation on Japanese dataset}
\label{sec:japanese}

We further evaluate DYNAC on a Japanese dataset containing the Corpus of Spontaneous Japanese (581 hours)~\cite{csj}, 181 hours of speech from a database developed by the Advanced Telecommunications Research Institute International~\cite{KUREMATSU1990357}, and 93 hours of our in-house recordings, 
with a static vocabulary size $K$ of 3,613.
Table \ref{castable} shows the results when $N$ = 203 phrases are registered in the bias list. 
Similar to the experiments on the LibriSpeech corpus (Table \ref{main_table}), DYNAC significantly reduces RTF compared to AR-CB methods \cite{sudo2024contextualized} while substantially improving B-WER. This demonstrates the effectiveness of DYNAC across languages with entirely different vocabulary structures.

\begin{table}[t]
\caption{Experimental results obtained on Japanese dataset. \textbf{Bold} values represent better results in the same category.}
\vspace*{-6.5mm}
\label{castable}
\begin{center}
\resizebox {0.9\linewidth} {!} {
\begin{tabular}{@{}lcccc}
\hline
Model & CER & U-CER & B-CER & RTF \\
\hline
\multicolumn{3}{l}{\textbf{\textit{Autoregressive}}} \\
CTC/attention & 9.9 & \textbf{8.2} & 21.8 & 0.171 \\
Dynamic vocab \cite{sudo2024contextualized} & \textbf{9.0} & 8.9 & \textbf{9.7} & \textbf{0.164}\\
\hline
\multicolumn{3}{l}{\textbf{\textit{Non-autoregressive}}} \\
Self-cond. CTC \cite{Nozaki2021RelaxingTC} & 11.8 & \textbf{9.9} & 25.9 & \textbf{0.023}\\
DYNAC (ours) & \textbf{10.6} & 10.6 & \textbf{10.6} & \textbf{0.023}\\
\hline
\end{tabular}
}
\end{center}
\vspace*{-5mm}
\end{table}

\section{Conclusion}

This paper proposes DYNAC (Dynamic Vocabulary-based NAR Contextualization), an NAR-CB method that integrates dynamic vocabulary into intermediate encoder layers using self-conditioned CTC, enabling efficient inference with low RTF. 
Experimental results demonstrate that DYNAC significantly reduces RTF while maintaining WER comparable to AR-CB methods on both the LibriSpeech 960 corpus and our in-house Japanese dataset.

\clearpage

\bibliographystyle{IEEEtran}
\bibliography{mybib}

\end{document}